%% file: main.tex
\documentclass[10pt,twocolumn,letterpaper]{article}

\usepackage[pagenumbers]{cvpr} 

\usepackage{graphicx}
\usepackage{amsmath}
\usepackage{amssymb}
\usepackage{booktabs}

\usepackage{mathtools}

\usepackage{multirow}

\usepackage[dvipsnames]{xcolor}
\usepackage{booktabs}
\usepackage{tabularx}
\usepackage{varwidth}
\usepackage{textcomp} 

\usepackage{floatrow}
\newfloatcommand{capbtabbox}{table}[][\FBwidth]
\usepackage{pifont}
\newcommand{\cmark}{\ding{51}}%
\newcommand{\xmark}{\ding{55}}%

\usepackage{dsfont}
\usepackage{bbm}

\usepackage{graphicx}
\usepackage{duckuments}
\usepackage{colortbl}

\usepackage{marvosym}

\newcommand{\Fref}[1]{Fig.~\ref{#1}}
\newcommand{\Sref}[1]{Sec.~\ref{#1}}
\newcommand{\Tref}[1]{Table~\ref{#1}}

\newcommand{\R}{\mathbb{R}}

\newcommand*\colourcheck[1]{%
  \expandafter\newcommand\csname #1check\endcsname{\textcolor{#1}{\ding{51}}}%
}
\colourcheck{ForestGreen}

\newcommand*\colourx[1]{%
  \expandafter\newcommand\csname #1x\endcsname{\textcolor{#1}{\ding{55}}}%
}
\colourx{blue}
\colourx{green}
\colourx{red}

\usepackage[pagebackref,breaklinks,colorlinks,citecolor=citecolor]{hyperref}

\definecolor{citecolor}{HTML}{0071BC}
\definecolor{linkcolor}{HTML}{ED1C24}

\usepackage[capitalize]{cleveref}
\crefname{section}{Sec.}{Secs.}
\Crefname{section}{Section}{Sections}
\Crefname{table}{Table}{Tables}
\crefname{table}{Tab.}{Tabs.}

\def\cvprPaperID{351} 
\def\confName{CVPR}
\def\confYear{2023}

\begin{document}

\title{Decomposed Cross-modal Distillation for RGB-based Temporal Action Detection}

\author{Pilhyeon Lee\textsuperscript{\rm 1} \quad Taeoh Kim\textsuperscript{\rm 2} \quad Minho Shim\textsuperscript{\rm 2} \quad Dongyoon Wee\textsuperscript{\rm 2} \quad
    Hyeran Byun\textsuperscript{\rm 1}\thanks{Corresponding author} \vspace{.5em} \\
\textsuperscript{\rm 1}Yonsei University \qquad
\textsuperscript{\rm 2}Naver Cloud, AI Tech.\\
{\tt\small \Letter \thinspace \thinspace lph1114@yonsei.ac.kr}
}

\maketitle

\begin{abstract}
Temporal action detection aims to predict the time intervals and the classes of action instances in the video.
Despite the promising performance, existing two-stream models exhibit slow inference speed due to their reliance on computationally expensive optical flow.
In this paper, we introduce a decomposed cross-modal distillation framework to build a strong RGB-based detector by transferring knowledge of the motion modality. 
Specifically, instead of direct distillation, we propose to separately learn RGB and motion representations, which are in turn combined to perform action localization. 
The dual-branch design and the asymmetric training objectives enable effective motion knowledge transfer while preserving RGB information intact. 
In addition, we introduce a local attentive fusion to better exploit the multimodal complementarity.
It is designed to preserve the local discriminability of the features that is important for action localization. 
Extensive experiments on the benchmarks verify the effectiveness of the proposed method in enhancing RGB-based action detectors.
Notably, our framework is agnostic to backbones and detection heads, bringing consistent gains across different model combinations.
\end{abstract}

\section{Introduction}
\label{sec:intro}

With the popularization of mobile devices, a significant number of videos are generated, uploaded, and shared every single day through various online platforms such as YouTube and TikTok.
Accordingly, there arises the importance of automatically analyzing untrimmed videos.
As one of the major tasks, temporal action detection (or localization)~\cite{shou2016temporal} has attracted much attention, whose goal is to find the time intervals of action instances in the given video.
In recent years, a lot of efforts have been devoted to improving the action detection performance~\cite{lin2018bsn,xu2020g-tad,Lin2019BMNBN,zhao2020bottom-up,zeng2019p-gcn,lee2020background,lee2021points,lee2021um}.

Most existing action detectors take as input two-stream data consisting of RGB frames and motion cues, \eg, optical flow~\cite{horn1981optical-flow,wedel2009improved,zach2007optical-flow}.
Indeed, it is widely known that different modalities provide complementary information~\cite{simonyan2014two,carreira2017quo,kazakos2019epic-fusion,xiao2020audiovisual-slowfast}.
To examine how much two-stream action detectors rely on the motion modality, we conduct an ablative study using a set of representative models\footnote{Each model is reproduced by its official codebase.}.
As shown in \Tref{table:motivation}, regardless of the framework types, all the models experience sharp performance drops when the motion modality is absent, probably due to the static bias of video models~\cite{li2018resound,li2019repair,choi2019cant-i-dance,kowal2022dive-static-dynamic}.
This indicates that explicit motion cues are essential for accurate action detection.

\input{figures/intro.tex}
\input{tables/motivation.tex}

However, two-stream action detectors impose a cycle of dilemmas for real-world applications due to the heavy computational cost of motion modality.
For instance, the most popular form of motion cues for action detection, TV-$L^1$ optical flow~\cite{wedel2009improved}, is not real-time, taking 1.8 minutes to process a 1-min $224\times224$ video of 30~fps on a single GPU~\cite{simonyan2014two}.
Although cheaper motion clues such as temporal gradient~\cite{zhao2018action-tg,wang2016temporal,xiao2022semi-action-tg} can be alternatives, two-stream models still exhibit inefficiency at inference by doubling the network forwarding process.
Therefore, it would be desirable to build strong RGB-based action detectors that can bridge the performance gap with two-stream methods.

To this end, we focus on cross-modal knowledge distillation~\cite{crasto2019mars,garcia2018modality-distillation}, where the helpful knowledge of motion modality is transferred to an RGB-based action detector during training in order to improve its performance.
In contrast to conventional knowledge distillation~\cite{hinton2015distillation,romero2014fitnets,zagoruyko2016attention-transfer,heo2019comprehensive} where the superior teacher guides the weak student, cross-modal distillation requires exploiting the complementarity of the teacher and student.
However, existing cross-modal distillation approaches~\cite{crasto2019mars,dai2021augmented-rgb} fail to consider the difference and directly transfer the motion knowledge to the RGB model (\Fref{fig:intro}a), as conventional distillation does.
By design, the RGB and motion information are entangled with each other, making it difficult to balance between them.
As a result, they often achieve limited gains without careful tuning.

To tackle the issue, we introduce a novel framework, named \textit{decomposed cross-modal distillation} (\Fref{fig:intro}b).
In detail, our model adopts the split-and-merge paradigm, where the high-level features are decomposed into appearance and motion components within a dual-branch design.
Then only the motion branch receives the distillation signal, while the other branch remains intact to learn appearance information.
For explicit decomposition, we adopt the shared detection head and the asymmetric objective functions for the branches.
Moreover, we design a novel attentive fusion to effectively combine the multimodal information provided by the two branches.
In contrast to existing attention methods, the proposed fusion preserves local sensitivity which is important for accurate action detection.
With these key components, we build a strong action detector that produces precise action predictions given only RGB frames.

We conduct extensive experiments on the popular benchmarks, THUMOS'14~\cite{THUMOS14} and ActivityNet1.3~\cite{caba2015activitynet}.
Experimental results show that the proposed framework enables effective cross-modal distillation by separating the RGB and motion features.
Consequently, our model largely improves the performance of RGB-based action detectors, exhibiting its superiority over conventional distillation.
The resulting RGB-based action detectors effectively bridge the gap with two-stream models.
Moreover, we validate our approach by utilizing another motion clue, \ie, temporal gradient, which has been underexplored for action detection.

To summarize, our contributions are three-fold:
1) We propose a decomposed cross-modal distillation framework, where motion knowledge is transferred in a separate way such that appearance information is not harmed.
2) We design a novel attentive fusion method that is able to exploit the complementarity of two modalities while sustaining the local discriminability of features.
3) Our method is generalizable to various backbones and detection heads, showing consistent improvements.

\section{Related Works}
\subsection{Temporal Action Detection}
Temporal action detection requires predicting temporal intervals as well as action categories for all action instances occurring in the video.
Conventional methods adopt the two-stage pipeline~(\ie, proposal-and-classification) and generate proposals by either sliding windows~\cite{chao2018rethinking,shou2017cdc,xiong2017pursuit,yang2018exploring,yuan2016temporal,zhao2017temporal} or predicting the per-frame starting and ending probabilities ~\cite{lin2020fast,Lin2019BMNBN,lin2018bsn,zhao2020bottom-up}.
Besides, several methods focus on proposal refinement to improve the detection performance~\cite{zeng2019p-gcn,qing2021lgte,zhu2021enriching}.
Meanwhile, analogous to one-stage object detection~\cite{liu2016ssd,redmon2016yolo}, anchor-free models are proposed for efficient action detection~\cite{lin2021afsd,zhang2022actionformer}.
Inspired by the recent success of DETR~\cite{carion2020detr}, query-based action detectors are also designed to streamline the complicated detection pipeline~\cite{tan2021relaxed,liu2022tadtr,shi2022react}.
In an orthogonal direction, some works showcase the benefit of end-to-end training of video backbones for temporal action detection~\cite{xu2021low-fidelity,liu2022e2e-tad,cheng2022tallformer}.

Most of the current action detectors leverage two-stream inputs for accurate action localization.
However, the optical flow takes heavy computations, suggesting the necessity of RGB-based action detection models.
In this paper, we propose a novel distillation framework to build a strong RGB-based action detector by transferring motion knowledge.

\subsection{Cross-modal Knowledge Distillation}
Knowledge distillation~\cite{hinton2015distillation} is originally devised to transfer the knowledge of large-scale models (teachers) to smaller ones (students).
Existing approaches can be categorized into three groups based on what types of knowledge are distilled: responses, features, and relations.
Response-based methods encourage the student to produce similar predictions to those of the teacher~\cite{hinton2015distillation,chen2017detection-distill,zhao2022decoupled-distill,beyer2022patient-consistent}.
Meanwhile, feature-based distillation methods pursue the matching of intermediate features between students and teachers~\cite{romero2014fitnets,zagoruyko2016attention-transfer,wang2019fine-grained-distill,shu2021channel-distill}.
Lastly, relation-based models focus on aligning the inter-sample relationship between the teacher and the student~\cite{liu2019graph-distill,park2019relational-distill,tian2019contrastive-distill}.
In practice, the three types of distillation are often used in a combinational form~\cite{zhao2022decoupled-distill,zheng2022localize-distill,dai2021general-instance}.

As a subset, cross-modal knowledge distillation aims to transfer knowledge of one modality to another modality.
Generally, the model taking the superior modality on the target task is selected as the teacher and guides the inferior modality model~\cite{ren2021learning-from-master,zhao2020knowledge-as-priors}.
On the other hand, in the regime of action analysis, there is no dominant modality, and it is crucial to grasp the complementarity of different modalities~\cite{simonyan2014two,carreira2017quo}.
However, existing works~\cite{crasto2019mars,dai2021augmented-rgb,garcia2018modality-distillation,luo2018graph-cross-distill} ignore this property and directly inject motion knowledge into the RGB model, leading to entangled representations.
In contrast, we propose a decomposed distillation framework, where different modality information is separately learned and recomposed for effective action detection.

\section{Method}
\noindent\textbf{Problem formulation.}~~
The input of temporal action detection models is a video $V \in \R^{T \times 3 \times H \times W}$ consisting of a total of $T$ frames with the size of $H \times W$.
Here the video length $T$ can vary across input videos.
Since most of the action detectors are based on two-stream inputs, the optical flow maps are additionally extracted from each pair of consecutive RGB frames.
We denote the optical flow maps by $F \in \R^{T \times 2 \times H \times W}$, where each map $F_t$ is composed of two channels that respectively estimate the displacements of $x$ and $y$ axes.
During training, the input video is labeled with its corresponding annotation $\Psi=\{(\varphi_m, \mathbf{y}_m)\}_{m=1}^{M}$, where $\mathbf{y}_m$ is the category label of the $m$-th action instance and $\varphi_m=(t_{s_m}, t_{e_m})$ indicates its starting and ending timestamps.
Typically, the class label $\mathbf{y}_m \in \R^{C}$ is a one-hot vector, where $C$ indicates the number of categories.
In the test phase, a two-stream action detector localizes the action instances based on both $V$ and $F$.
Differently, to bypass time-consuming optical flow extraction and inefficient inference, we aim to build an RGB-based action detector that takes only $V$ as input.
To this end, we leverage a pre-trained motion model as the teacher and transfer its knowledge to the RGB-based action detector in a decomposed way.

\noindent\textbf{Motion modality.}~~
Conventionally, optical flow is utilized as the motion modality for action detection~\cite{chao2018rethinking,lin2018bsn}.
Meanwhile, another form of the motion modality, \ie, temporal gradient, has recently been explored for action recognition~\cite{xiong2021multiview-tg,xiao2022semi-action-tg}.
Inspired by this, we explore temporal gradient for temporal action detection for the first time.
The temporal gradient maps $G \in \R^{T \times 3 \times H \times W}$ is defined by the residual difference between two consecutive frames, \ie, $G_t = V_t - V_{t-1}$, which implicitly captures motion information such as camera moving, object moving, \textit{etc}.
\Fref{fig:example_frames} exemplifies an RGB frame and the corresponding optical flow and temporal gradient.
Although vulnerable to noises like environmental changes, the temporal gradient can be obtained on-the-fly and thus have the potential to be utilized as a weak form of motion modality.
Indeed, it is shown in the experiments that its motion knowledge helps to improve the action detection performance.
In the following, we will elaborate on our framework using optical flow maps $F$ as the motion modality by default, but they can be replaced by temporal gradient maps $G$ without loss of generality.

\subsection{Motion Teacher Training}
In this section, we describe the training of the motion teacher.
Given the input optical flow $F$, the video backbone extracts 1D spatially pooled features $z^{\text{mot}} \in \R^{T/r_T \times C}$, where $r_T$ indicates the temporal downsampling rate and $C$ is the channel dimension.
Here $r_T$ can vary depending on the backbone choices.
The extracted features go through the action detection head, resulting in a set of predicted action boundaries and scores $\hat{\Psi}^{\text{mot}}=\{(\hat{\varphi}_n^{\text{mot}}, \hat{\mathbf{y}}_n^{\text{mot}})\}_{n=1}^{N}$, where $\hat{\varphi}_n^{\text{mot}} = (\hat{t}_{s_n}^{\text{mot}}, \hat{t}_{e_n}^{\text{mot}})$ is the predicted action proposals, $N$ is the number of predictions, and $\hat{\mathbf{y}}_n$ is the class probability with Sigmoid activation.
We leave the model architecture and the detailed prediction process to be abstract since our framework can be applied to any framework type, \eg, anchor-based~\cite{xu2020g-tad,long2019gaussian}, anchor-free~\cite{lin2021afsd,zhang2022actionformer}, query-based~\cite{liu2022tadtr,tan2021relaxed}.
We show in experiments that our approach brings consistent improvements for different types of heads.
The action predictions are grouped by positive and negative sets based on pre-defined principles; some works perform thresholding the intersection-over-union~(IoU) with ground truths~\cite{chao2018rethinking,shou2017cdc}, while other query-based methods rely on the Hungarian matching~\cite{liu2022e2e-tad}.
We denote the positive and negative sets by $\hat{\Psi}_P^{\text{mot}}$ and $\hat{\Psi}_N^{\text{mot}}$, respectively.
The predictions in the positive set are labeled by the best-matched ground-truth intervals and classes $\{(\varphi_n, \mathbf{y}_n)\}_{n=1}^{|\hat{\Psi}_P^{\text{mot}}|}$.
On the other hand, the predictions in the negative set are annotated as the background class.
Despite minor differences, the training objective of action detectors generally consists of a classification loss, a regression loss, and a completeness loss.

\input{figures/example_frames.tex}

1) The classification loss is defined by a weighted sum of the focal binary cross-entropy function as follows.
\begin{equation}
\resizebox{1.0\columnwidth}{!}{
$
\begin{aligned}
    \mathcal{L}_{\text{cls}}=\frac{\alpha_{P}}{|\hat{\Psi}_P^{\text{mot}}|} \sum_{ \hat{\mathbf{y}}_{n} \in \hat{\Psi}_P^{\text{mot}} } \ell_{\text{focal}}(\hat{\mathbf{y}}_{n}, \mathbf{y}_n) +\frac{\alpha_{N}}{|\hat{\Psi}_N^{\text{mot}}|}\sum_{\hat{\mathbf{y}}_{n} \in \hat{\Psi}_N^{\text{mot}} } \ell_{\text{focal}}(\hat{\mathbf{y}}_n, {\mathbf{0}}),
\end{aligned}
$
}
\label{eq:loss_cls}
\end{equation}
where $\ell_{\text{focal}}(\hat{\mathbf{y}}, \mathbf{y})$ denotes the focal binary cross-entropy loss~\cite{lin2017focal}, $\alpha_{*}$ is the term for balancing between positive and negative samples, and $\mathbf{0} \in \R^{C}$ is the $C$-dimensional zero vector indicating the background class.

2) The regression loss is defined by the $L_1$-distance between ground truths and the predicted offsets as follows.
\begin{equation}
    \mathcal{L}_{\text{reg}} = \frac{1}{|\hat{\Psi}_P^{\text{mot}}|} \sum_{ \hat{\varphi}_{n} \in \hat{\Psi}_P^{\text{mot}} } \ell_{1}(\hat{\varphi}_{n}, \varphi_{n}),
\label{eq:loss_reg}
\end{equation}
where $\ell_{1}(\hat{\varphi}, \varphi)$ denotes the (smooth) $L_1$ distance between the two offsets. It is worth noting that the regression loss is computed only for the positive set.

3) The completeness loss is utilized to maximize the IoUs between ground truths and the predicted proposals.
\begin{equation}
    \mathcal{L}_{\text{comp}} = \frac{1}{|\hat{\Psi}_P^{\text{mot}}|} \sum_{ \hat{\varphi}_{n} \in \hat{\Psi}_P^{\text{mot}} } \big( 1 - \frac{|\hat{\varphi}_n \cap \varphi_n|}{|\hat{\varphi}_n \cup \varphi_n|} \big).
\label{eq:loss_comp}
\end{equation}
The completeness loss is defined only for the positive set.

The overall objective function is a weighted sum of the above losses: $\lambda_{\text{cls}}\mathcal{L}_{\text{cls}} + \lambda_{\text{reg}}\mathcal{L}_{\text{reg}} + \lambda_{\text{comp}}\mathcal{L}_{\text{comp}}$.
After training, the motion model $\mathcal{M}^{\text{mot}}$ is able to extract discriminative motion features and will serve as the teacher for distillation.

\subsection{Decomposed Cross-modal Distillation}
\paragraph{Motivation.}
Previous cross-modal knowledge distillation frameworks~\cite{crasto2019mars,dai2021augmented-rgb} transfer motion knowledge to the RGB model in a direct way, so that the RGB model is encouraged to produce similar predictions with the motion model.
Such an approach, however, would result in sub-optimal solutions, as the complementarity of the two modalities is not considered.
To handle the challenge, we design a novel decomposed distillation framework, where RGB and motion information are separately learned and later fused to effectively exploit their complementarity. 
The training process of our distillation framework is illustrated in \Fref{fig:architecture}a.

Similar to the motion model, given RGB frames $V$, the video backbone first produces the spatiotemporal features $z^{\text{RGB}} \in \R^{T/r_T \times C}$.
The representation $z^{\text{RGB}}$ is projected into two different spaces, \ie, motion and appearance spaces.
The projections are implemented by a series of 1D convolutional layers, and we denote this process by $\phi_{\text{app}}(z^{\text{RGB}}), \phi_{\text{mot}}(z^{\text{RGB}}) \in \R^{T/r_T \times D}$, respectively.
Afterward, the projected features serve as input to the subsequent separate branches to predict the temporal intervals of action instances.
In the following, we describe how the separated features learn their own representations.

\input{figures/architecture.tex}
\subsubsection{Motion branch}
The motion branch is supposed to learn motion information from the RGB frames.
For this purpose, the pre-trained motion model $\mathcal{M}^{\text{mot}}$ (teacher) explicitly guides the motion branch (student) by distilling its knowledge.
A variety of distillation approaches can be applied to our framework, and we adopt two representative methods in the following.

1) The response-based distillation loss~\cite{hinton2015distillation,chen2017detection-distill} utilizes the teacher's predictions as pseudo-ground truths and encourages the student to mimic the behavior of the teacher.
\begin{equation}
\begin{aligned}
    \mathcal{L}_{\text{respon}} &= \lambda_{\text{cls}}\mathcal{L}_{\text{respon}}^{\text{cls}} + \lambda_{\text{reg}}\mathcal{L}_{\text{respon}}^{\text{reg}}, \\
    \text{where}~~
    \mathcal{L}_{\text{respon}}^{\text{cls}}&=\frac{1}{|\hat{\Psi}|} \sum_{ n=1 }^{|\hat{\Psi}|} \ell_{\text{focal}}(\hat{\mathbf{y}}_{n}, \hat{\mathbf{y}}_n^{\text{mot}})\\
    \text{and}~~
    \mathcal{L}_{\text{respon}}^{\text{reg}} &= \frac{1}{|\hat{\Psi}|} \sum_{ n=1 }^{|\hat{\Psi}|} \ell_{1}(\hat{\varphi}_{n}, \hat{\varphi}_{n}^{\text{mot}}).
\label{eq:response_distill}
\end{aligned}
\end{equation}
where $(\hat{\varphi}_{n}^{\text{mot}}, \hat{\mathbf{y}}_n^{\text{mot}})$ is the corresponding teacher's predictions of the student's ones.
Here $\ell_{\text{focal}}$ and $\ell_1$ are defined as same in Eq.~\ref{eq:loss_cls} and \ref{eq:loss_reg}, respectively.

2) The feature-based distillation loss~\cite{romero2014fitnets} encourages the matching of the intermediate features from the student and the teacher, which is formulated as:
\begin{equation}
    \mathcal{L}_{\text{feat}} = \frac{1}{T}\sum_{t=1}^{T}\lVert \phi_{\text{mot}}(z^{RGB}_t) - z^{\text{mot}}_t \rVert_2^2,
\label{eq:feature_distill}
\end{equation}
where the projection layer $\phi_{\text{mot}}(\cdot)$ can be viewed as an adaptation layer~\cite{chen2022simkd} that helps to ease the feature matching between different modality features.
Note that although this equation represents the matching of backbone features, other features from any intermediate layer can be aligned.

To summarize, the total training objective function of the motion branch is defined by the weighted sum of the distillation losses: $\mathcal{L}_{\text{distill}} = \mathcal{L}_{\text{respon}} + \lambda_{\text{feat}}\mathcal{L}_{\text{feat}}$.
With the guidance of the pre-trained motion model, the motion branch learns motion representations given RGB frames as inputs.
Note that our framework is general and can utilize other improved distillation methods, but it is beyond the scope of this paper.

\subsubsection{Appearance branch}
On the other side, to prevent the model from losing rather static RGB information during distillation, we train the appearance branch jointly with the motion branch.
The training objective is the same with the motion teacher $\mathcal{M}^{\text{mot}}$, except that the input is RGB frames.
Here we define the training objective of the appearance branch as: 
$\mathcal{L}_{\text{app}} = \lambda_{\text{cls}}\mathcal{L}_{\text{cls}} + \lambda_{\text{reg}}\mathcal{L}_{\text{reg}} + \lambda_{\text{comp}}\mathcal{L}_{\text{comp}}$.
Being trained with the conventional detection loss, the appearance branch is able to keep the appearance information intact during distillation which is useful for action detection.

\subsubsection{Discussion}
Our decomposed distillation model contains two separate branches with different design purposes.
Specifically, the motion branch needs to imitate the behavior of the motion teacher, while the appearance branch is supposed to learn the original appearance information from the same RGB inputs.
To achieve this goal, we propose to endow the branches with conflicting training objectives.
However, the two branches may reach a degenerate solution, learning similar representations without distinction and relying solely on the powerful detection head to produce different predictions.
To tackle the potential issue, we make the branches share the weights of the classifier and the regressor in the detection heads.
With these two key designs, we can lead the two branches to learn the different information of RGB and motion modalities, thus achieving their complementarity of theirs in the subsequent fusion stage.

\subsection{Local Attentive Fusion}
Provided that the appearance and motion features have learned the expected information, the next question is how to fuse the different information from them.
An intuitive way would be to perform concatenation before feeding them to the joint detection head, also known as early fusion~\cite{joze2020mmtm}.
However, we argue that such a naive way does not help much since erroneous predictions of one modality can propagate to the other.
Instead, we utilize an attentive fusion mechanism to maximize the harmonizing effect of the two different modalities.

Our motivation is that one modality can enhance the other by highlighting the agreement between different modalities.
One can try applying cross-attention for that purpose, where features of one modality aggregate information from those of the other modality based on feature similarity.
Although the cross-attention proves to be effective in various fields~\cite{gorti2022x-pool,wei2020multi}, we experimentally find that it does not help much to improve the action detection performance.
We conjecture that this is because the cross-attention hampers local discriminability of the features by gathering information along the temporal dimension, \ie, feature over-smoothing, which is also observed in the literature~\cite{tan2021relaxed,gong2021vit-oversmoothing}.
To bypass the issue, we propose a new attentive fusion that sustains locally discriminative features for action detection.

Let us call the modality to be enhanced the target modality and the other the reference modality.
At first, we build a single representative feature for the target modality by aggregating the whole features along the temporal dimension, \ie, $f^{\text{target}}=\psi (f^{\text{target}}_1, \dots,f^{\text{target}}_T) \in \R^{D}$, where the aggregation function can be implemented by various pooling methods.
Thereafter, it serves as the query and the individual features from the assistant modality are deemed as the key.
Then, the query-key matching process is defined by:
\begin{equation}
    \omega_t = \sigma\big[(W_{\text{query}}^{\top}f^{\text{target}}) \odot (W_{\text{key}}^{\top}f_t^{\text{ref}})\big],
\end{equation}
where $W_{\text{query}}, W_{\text{key}} \in \R^{D \times D}$ respectively indicate the projection matrices for the query and keys, $\odot$ denotes the Hadamard product, and $\sigma(\cdot)$ is the sigmoid activation.
Here the resulting weight $\omega_t \in [0, 1]^D$ can be viewed as the channel attention weights that are derived from the similarity between the two modalities.
Therefore, we can suppress erroneous predictions of the target modality and emphasize the mutually agreed information by applying the weight to the features of the target modality, \ie, $\tilde{f}_t^{\text{target}} = \omega_t \cdot f_t^{\text{target}}$.
The overall process of our attentive fusion is illustrated in \Fref{fig:fusion}.
Importantly, our attentive fusion does not fuse information of different temporal points and thus can preserve the local sensitivity of the features.
The advantage of our local attentive fusion over the conventional cross-attention will be verified in \Sref{sec:experiments}.
After enhancing the appearance and motion features based on each other, we concatenate and put them into the joint detection head.
The detection head is trained to perform the action detection given the enhanced multimodal features.
The loss function of the detection head is the same as that of the appearance branch: $\mathcal{L}_{\text{joint}} = \lambda_{\text{cls}}\mathcal{L}_{\text{cls}} + \lambda_{\text{reg}}\mathcal{L}_{\text{reg}} + \lambda_{\text{comp}}\mathcal{L}_{\text{comp}}$.

\input{figures/fusion.tex}

\subsection{Joint Training and Inference}
Our framework is trained in an end-to-end manner.
The total training objective of our model is defined as $\mathcal{L}_{\text{total}} = \mathcal{L}_{\text{app}} + \mathcal{L}_{\text{distill}} + \mathcal{L}_{\text{joint}}$.
At inference, we discard the detection heads of the branches and predict action intervals and classes using the joint head, as depicted in \Fref{fig:architecture}b.
It is noteworthy that our model takes only RGB frames as input and localizes action instances based on multimodal information.

\section{Experiments}
\label{sec:experiments}

\subsection{Experimental Setups}
\noindent\textbf{Datasets.}~~
We evaluate our framework on the two most popular benchmarks for temporal action detection: THUMOS'14~\cite{THUMOS14} and ActivityNet1.3~\cite{caba2015activitynet}.
THUMOS'14 consists of 200 and 213 videos respectively for training and testing with 20 action categories.
It is a challenging dataset since it contains frequent action instances, \eg, 15 instances per video on average.
Therefore, we utilize it as the main dataset for experiments.
On the other hand, ActivityNet1.3 is a relatively large-scale dataset containing 10,024, 4,926, and 5,044 videos respectively for training, validation, and testing.
As the ground truths of the test set are unavailable, we evaluate the comparative models on the validation set.

\noindent\textbf{Evaluation metrics.}~~
The standard protocol for evaluating action detectors is mean average precisions (mAPs) at different intersection-over-union (IoU) thresholds.
Following the convention~\cite{chao2018rethinking,xu2020g-tad,liu2021muses}, we set the thresholds to [0.3:0.7] with a step size of 0.1 for THUMOS'14 and [0.5:0.95] with a step size of 0.05 for ActivityNet1.3.

\input{tables/ablation_components.tex}

\noindent\textbf{Implementation details.}~~
Our framework is agnostic to the choices of backbones and action detection heads.
To confirm the generalizability of the proposed method, we conduct experiments using different combinations.
The video backbones are TSM18~\cite{lin2019tsm}, TSM50, I3D~\cite{carreira2017quo}, and Slowfast50~\cite{feichtenhofer2019slowfast}.
For the detection head, we utilize the representative models of three different detector types: anchor-based (GTAD~\cite{xu2020g-tad}), anchor-free (Actionformer~\cite{zhang2022actionformer}), and query-based (TadTR~\cite{liu2022tadtr}).
For model implementation and hyperparameter settings, we strictly follow the official codebases.
All the backbones are pre-trained on Kinetics-400~\cite{carreira2017quo}.
Following the previous works~\cite{lin2018bsn,lin2021afsd,zhang2022actionformer}, we use the TV-$L^1$ algorithm~\cite{Andrews2002SupportVM} to extract dense optical flow.
Since only the I3D model has pre-trained weights on the optical flow, we utilize it as the teacher network in the distillation setting of optical flow.
Besides, we also conduct experiments using the temporal gradient as the motion modality, where the identical backbone networks serve as the teacher to guide the RGB model.
We follow Liu~\etal~\cite{liu2022e2e-tad} for data processing, where the input is a video with the size of 96$\times$96 and the length of 25.6 seconds.
Our model is trained in an end-to-end fashion using the Adam optimizer~\cite{kingma2014adam} with the learning rate of 1e-4 for 20 epochs.

\input{tables/ablation_fusion.tex}
\input{figures/qualitative_results.tex}
\subsection{Analysis}
\subsubsection{Effect of each component}
To analyze the impact of each component, we conduct an ablation study on THUMOS'14 in \Tref{tab:ablation_components}.
For this study, we utilize TSM18~\cite{lin2019tsm} and Actionformer~\cite{zhang2022actionformer} as the backbone and the detection head, respectively. 
We first set our baseline using the pure RGB-based action detector, whose average mAP is 43.6~\%.
On top of it, we apply two different types of distillation with temporal gradient as the motion modality, namely the conventional and decomposed ones.
For the decomposed distillation, we simply concatenate features from the two branches and feed them to the joint branch for action detection.
As a result, the conventional distillation achieves a limited performance gain, while our decomposed distillation greatly boosts the detection performance~(2$^\text{nd}$-3$^\text{rd}$ rows).
This indicates that the way of directly transferring the motion knowledge to the RGB model is prone to a sub-optimal solution due to entangled information.
When adopting the proposed local attentive fusion, our model better grasps the multimodal complementarity and shows a further performance gain of 1.4~\%~(4$^\text{th}$ row), indicating the importance of information fusion.
The resulting distilled RGB model improves the baseline by 3.0~\%, which clearly verifies the effectiveness of the proposed methods.

\subsubsection{Ablation study on fusion}
To verify the effectiveness of the proposed local attentive fusion, we compare it with a variety of fusion methods.
In specific, we employ two naïve fusion methods, \ie, concatenation and summation, and three attention-based approaches~\cite{vaswani2017attention,yang2022background-constraint}, \ie, self-, cross-, and difference-attention.
The results are shown in \Tref{tab:ablation_fusion}, where we observe that the naive fusion shows decent performance, while the attention-based methods perform even poorer than the naïve ones.
We conjecture that this is due to the over-smoothing of the features, which are known to be a side effect of attention-based methods~\cite{gong2021vit-oversmoothing,shi2022bert-oversmoothing,tan2021relaxed}, leading to less discriminability in action boundary detection.
On the contrary, the proposed local attentive fusion successfully preserves the local sensitivity of features during multimodal information exchange, thereby achieving the best action localization performance.

\input{tables/experiments_backbones.tex}

\subsubsection{Generalizability}
It is worth noting that our distillation framework is general and can be applied to any video backbones and detection heads.
To analyze the generalizability of our method, we conduct comprehensive experiments on THUMOS'14.

The experimental results on four different video backbones are presented in \Tref{tab:backbones}.
We use Actionformer~\cite{zhang2022actionformer} as the detection head for this experiment.
It can be noticed that regardless of the backbone choices, our decomposed distillation consistently improves the performance with large gains.
In addition, even the strong backbone, \ie, Slowfast, benefits from the motion knowledge distillation, showing a performance gain of 3.0~\% when using optical flow as the teacher modality.
In addition, the results using different detection heads are provided in \Tref{tab:heads}, where TSM18~\cite{lin2019tsm} is employed as the backbone network.
Again, it is noticeable that all the heads benefit from the decomposed distillation to a large extent.
In detail, they show significant performance boosts when being distilled the motion knowledge from the optical flow.
On the other hand, albeit being efficiently obtained from RGB frames, the temporal gradient also brings nontrivial performance gains, shedding light on its potential to serve as an efficient motion modality.
To summarize, these experiments clearly validate the generalizability of the proposed distillation framework.

\input{tables/experiments_heads.tex}

\input{tables/quantitative_comparison.tex}

\subsection{Qualitative Results}
To analyze where the performance gains come from, we provide several qualitative comparisons in \Fref{fig:qualitative_results}.
Specifically, we visualize the detection results from the RGB baseline, the motion teacher, and the resultant RGB-based model of our decomposed distillation.
In both examples, it can be observed that the RGB-based baseline shows unsatisfying performance, indicating the importance of motion modality.
Meanwhile, the motion teacher suffers from the static video with tiny movements~(\Fref{fig:qualitative_results}b), resulting in inaccurate localization results.
On the other hand, our model produces precise action proposals for both examples by successfully fusing the multimodal information learned within the decomposed distillation framework.

\subsection{Comparison with State-of-the-arts}
The state-of-the-art comparison of THUMOS'14 and ActivityNet1.3 is shown in \Tref{tab:quantitative_comparison}.
For the comparison, we utilize Slowfast50~\cite{feichtenhofer2019slowfast} and Actionformer~\cite{zhang2022actionformer} for the backbone and the detection head, respectively.
To make the comparison clear, we separate the entries based on whether the models take as input two-stream data or not.
The first thing we can observe is that the RGB-based approaches fall largely behind the two-stream approaches, especially on THUMOS'14 compared to ActivityNet1.3.
This is because the THUMOS'14 benchmark contains frequent actions occurring, \eg, 15 instances per video on average, and therefore the motion modality has a large impact on localizing the actions.
On the other hand, ActivityNet1.3 have sparse and relatively long action instances, \eg, 1.5 instances per video on average, and it is widely known that classification rather than localization is important for the dataset due to a large number of action classes.
In the comparison results on THUMOS'14, our model achieves the state-of-the-art performance among the RGB-based action detectors.
In addition, it surpasses many two-stream action detection models even without relying on the motion modality during inference.
This signifies the efficacy of our distillation framework that enables simulating two-stream predictions by effectively distilling the motion knowledge and exploiting the multimodal complementarity.
On ActivityNet1.3, our model also sets a new state-of-the-art among the RGB-based detectors and shows comparable performance with two-stream approaches.

\section{Conclusion}
In this paper, we have presented a new paradigm for cross-modal distillation.
Specifically, we pointed out that existing distillation approaches inevitably entangle the RGB and motion information during the distilling process.
To handle the issue, we propose a decomposed distillation pipeline that enables separate learning of different modalities.
Furthermore, we design a local attentive fusion to sustain the local discriminability of features during integrating multimodal information, thereby accomplishing accurate action detection.
In the extensive experiments, we verified the effectiveness of our distillation framework and local attentive fusion.
Moreover, our model successfully bridges the gap between two-stream and RGB-based action detectors while preserving efficiency at test time.
Notably, our framework is generalizable to various combinations of video backbones and action detection heads, demonstrating consistent performance improvements.
In the future, it would be interesting to explore our decomposed distillation framework for other multimodal tasks.

\section*{Acknowledgements}
{
\small
\noindent This project was partly supported by the NAVER Corporation and the National Research Foundation of Korea grant funded by the Korea government (MSIT) (No. 2022R1A2B5B02001467).
Part of the computational work was conducted on NAVER Smart Machine Learning (NSML) platform~\cite{sung2017nsml,kim2018nsml}.

}

{\small
\bibliographystyle{ieee_fullname}
\bibliography{egbib}
}

\end{document}

%% file: figures/intro.tex
\begin{figure}[t]
    \centering
    \includegraphics[width=0.97\linewidth]{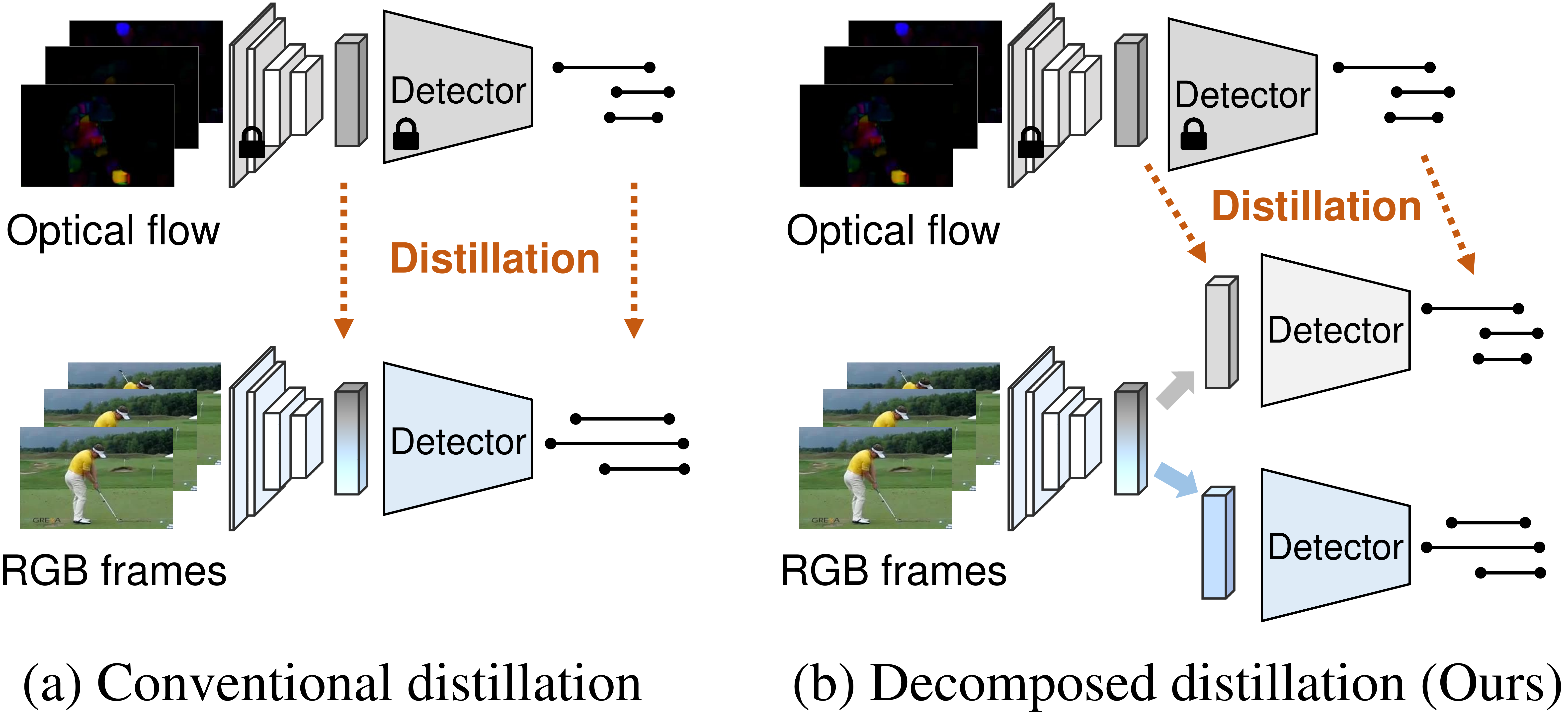}
    \caption{Comparison between conventional distillation and ours.}
    \label{fig:intro}
\end{figure}

%% file: tables/motivation.tex
\begin{table}[t]
\centering
\resizebox{0.8\columnwidth}{!}{
\begin{tabular}{c|l|ccc}
\toprule
\multirow{2}{*}{Framework} &
\multicolumn{1}{c|}{\multirow{2}{*}{Method}} &
\multicolumn{3}{c}{Average mAP (\%)} \\
& & RGB+OF & RGB & $\Delta$ \\
      \midrule\midrule
\multirow{1}{*}{Anchor-based}
        & G-TAD~\cite{xu2020g-tad}  & 41.5  & 26.9  & $-$14.6 \\
        \midrule
\multirow{2}{*}{Anchor-free}
       & AFSD~\cite{lin2021afsd}  & 52.4  & 43.3  & $-$9.1 \\
       & Actionformer~\cite{zhang2022actionformer}  & 62.2  & 55.5  & $-$6.7 \\
       \midrule
\multirow{1}{*}{DETR-like} 
       & TadTR~\cite{liu2022tadtr}  & 56.7  & 46.0  & $-$10.7 \\
    \midrule
\multirow{1}{*}{Proposal-free} 
       & TAGS~\cite{nag2022proposal-free}  & 52.8  & 47.9  & $-$4.9 \\
    \bottomrule
\end{tabular}
}
\caption{
Impact of motion modality. We measure the average mAP under the IoU thresholds of [0.3:0.7:0.1] on THUMOS'14.
}
\label{table:motivation}
\end{table}

%% file: figures/example_frames.tex
\begin{figure}[t]
    \centering
    \includegraphics[width=0.79\linewidth]{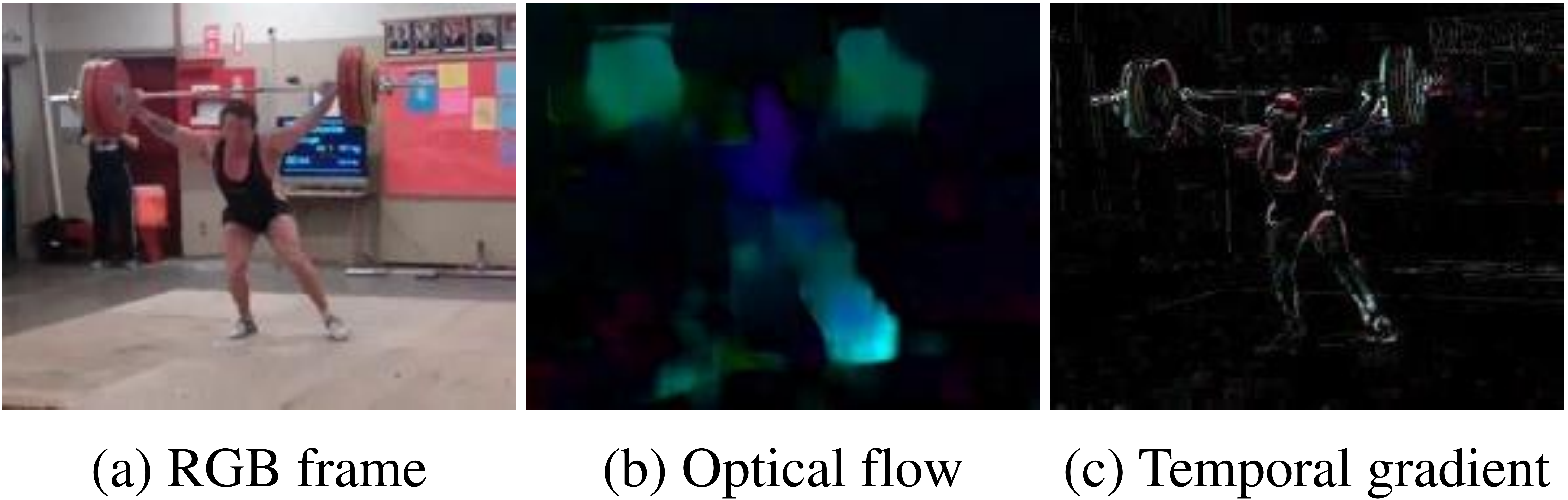}
    \caption{An example of an RGB frame and the corresponding optical flow map and temporal gradient map.}
    \label{fig:example_frames}
\end{figure}

%% file: figures/architecture.tex
\begin{figure}[t]
    \centering
    \includegraphics[width=0.98\columnwidth]{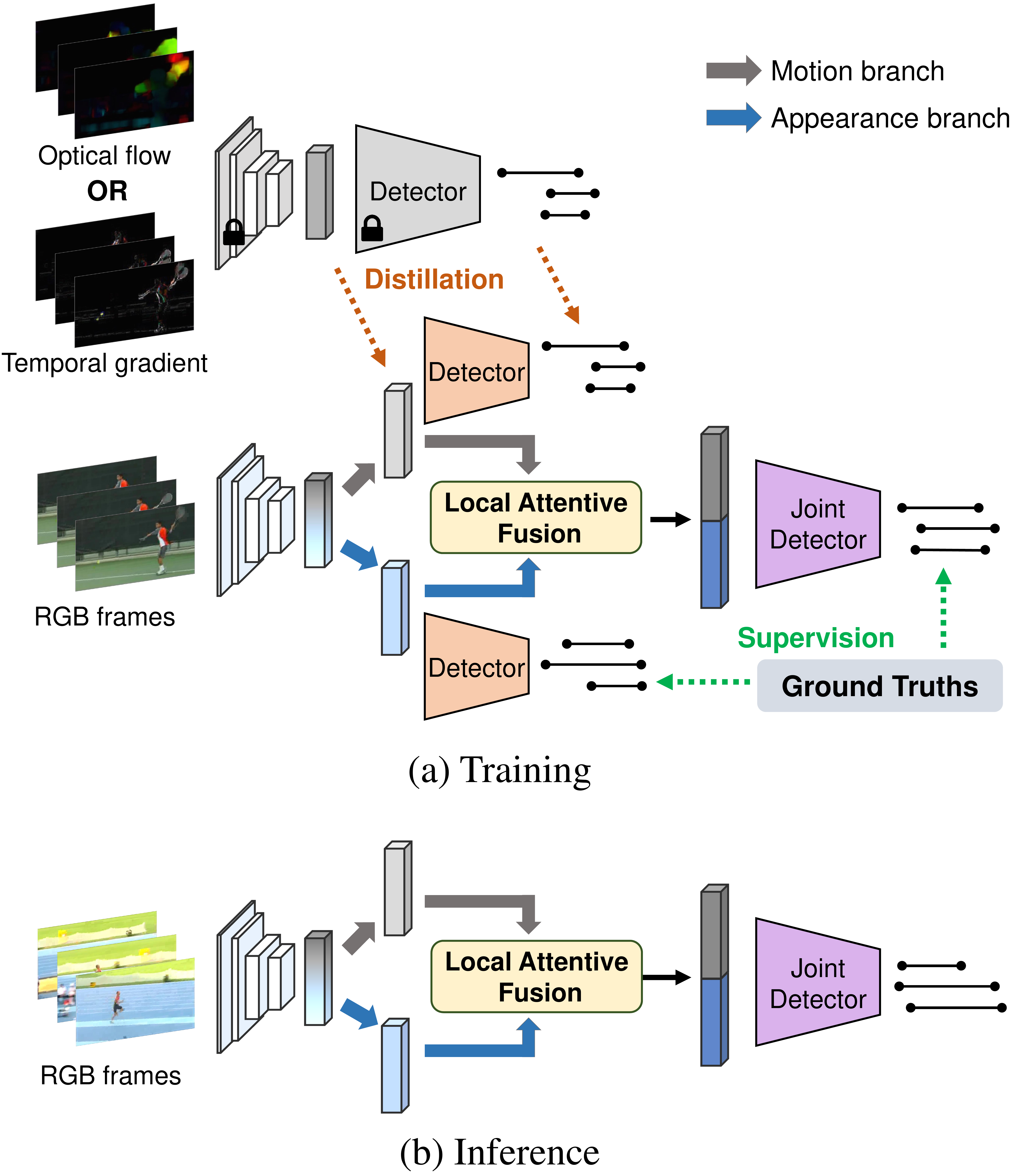}
    \caption{Overall workflow of the proposed framework. (a) During training, our model performs decomposed cross-modal distillation by explicitly separating motion and appearance branches. Here the detection heads of two branches are shared. (b) At inference, our model takes as input only RGB frames and precisely localizes action instances based on multimodal information.}
    \label{fig:architecture}
\end{figure}

%% file: figures/fusion.tex
\begin{figure}[t]
    \centering
    \includegraphics[width=0.98\linewidth]{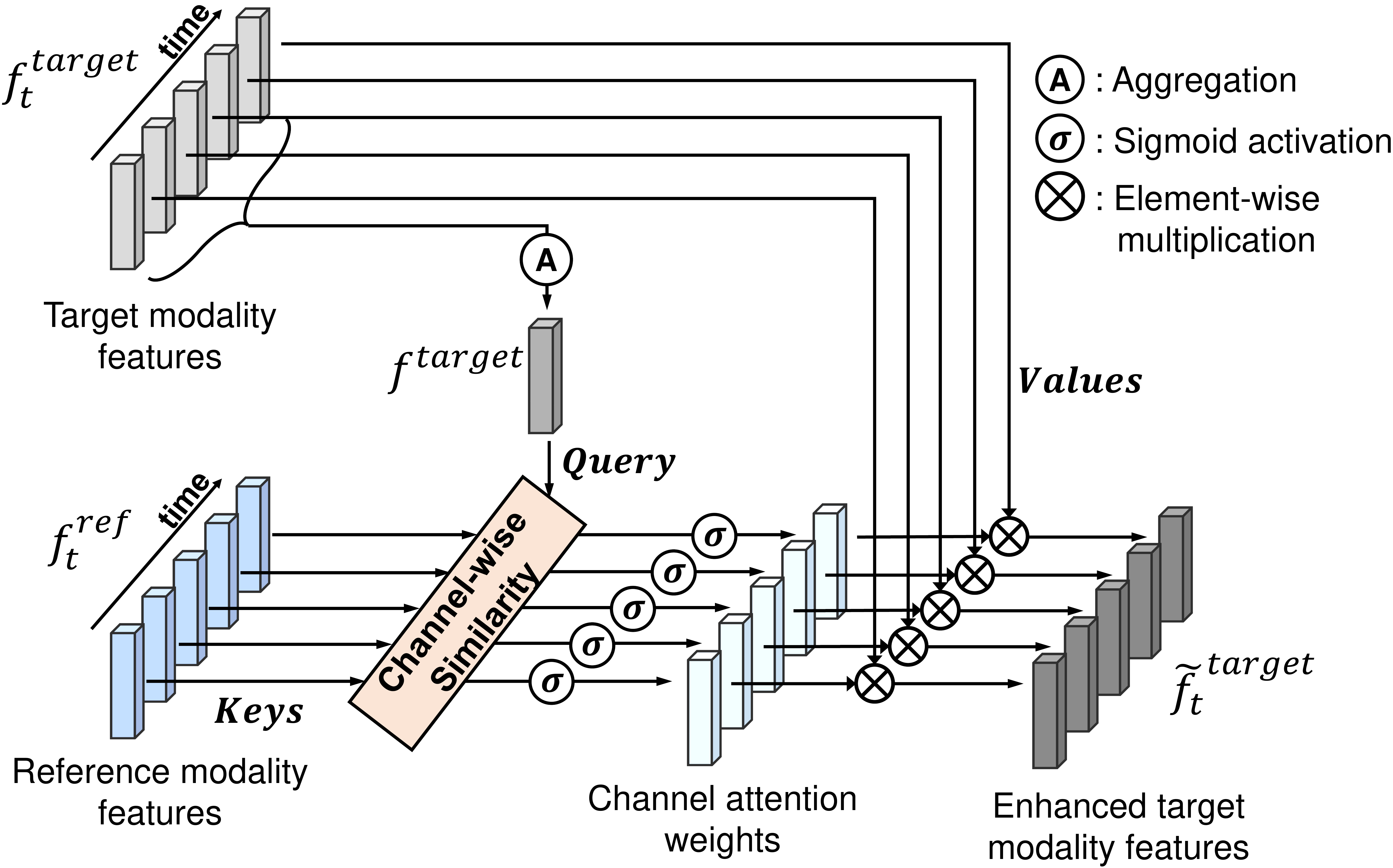}
    \caption{Illustration of the proposed local attentive fusion.}
    \label{fig:fusion}
\end{figure}

%% file: tables/ablation_components.tex
\begin{table}[t]
\centering
\resizebox{1.0\textwidth}{!}{
\begin{tabular}{cc|c|ccccc|c}
    \toprule
    \multicolumn{2}{c|}{distillation}  & \multirow{2}{*}{local attn.} & \multicolumn{5}{c|}{mAP@IoU (\%)}  & \multirow{2}{*}{AVG}  \\
     conven. & decomp.  & & 0.3  & 0.4  & 0.5 & 0.6 & 0.7  &   \\
    \midrule\midrule
    \xmark  & \xmark & \xmark & 62.3   & 55.2   & 46.2   & 33.8   & 20.4   & 43.6 \\
    \midrule
    \cmark  & & & 62.5  & 55.7  & 47.3  & 35.1  & 21.8  & 44.5   \\
       & \cmark  &    & 63.3 & 56.2  & 47.9  & 36.1  & 22.9  & 45.2  \\
       & \cmark   & \cmark & 64.4   & 58.0   & 49.0  & 37.5  & 24.1  & 46.6 \\
    \bottomrule
\end{tabular}
}
\caption{
Ablation study on the effect of each component. The comparative methods are evaluated on THUMOS'14.
}
\label{tab:ablation_components}
\end{table}

%% file: tables/ablation_fusion.tex
\begin{table}[t]
\centering
\resizebox{0.85\columnwidth}{!}{
\begin{tabular}{c|ccccc|c}
\toprule
\multirow{2}{*}{Fusion} &
\multicolumn{5}{c|}{mAP@IoU (\%)} & \multirow{2}{*}{AVG} \\
& 0.3 & 0.4 & 0.5 & 0.6 & 0.7 &  \\
      \midrule\midrule
concat. & 63.3 & 56.2  & 47.9  & 36.1  & 22.9  & 45.2  \\
sum.  & 62.6  & 56.1  & 47.5  & 36.1  & 23.0  & 45.1\\
self-attn.  & 63.8  & 56.3  & 46.7  & 34.2  & 21.9  & 44.6 \\
cross-attn.  & 63.1  & 54.5  & 46.4  & 35.4  & 21.7  & 44.2 \\
diff.-attn.  & 61.8  & 54.8  & 46.3  & 32.6  & 21.0  & 43.3 \\
local attn. (Ours)  & 64.4   & 58.0   & 49.0  & 37.5  & 24.1  & 46.6 \\
    \bottomrule
\end{tabular}
}
\caption{
Ablation study on the fusion mechanism. The comparative methods are evaluated on THUMOS'14.
}
\label{tab:ablation_fusion}
\end{table}

%% file: figures/qualitative_results.tex
\begin{figure*}[t]
    \centering
    \includegraphics[width=0.85\linewidth]{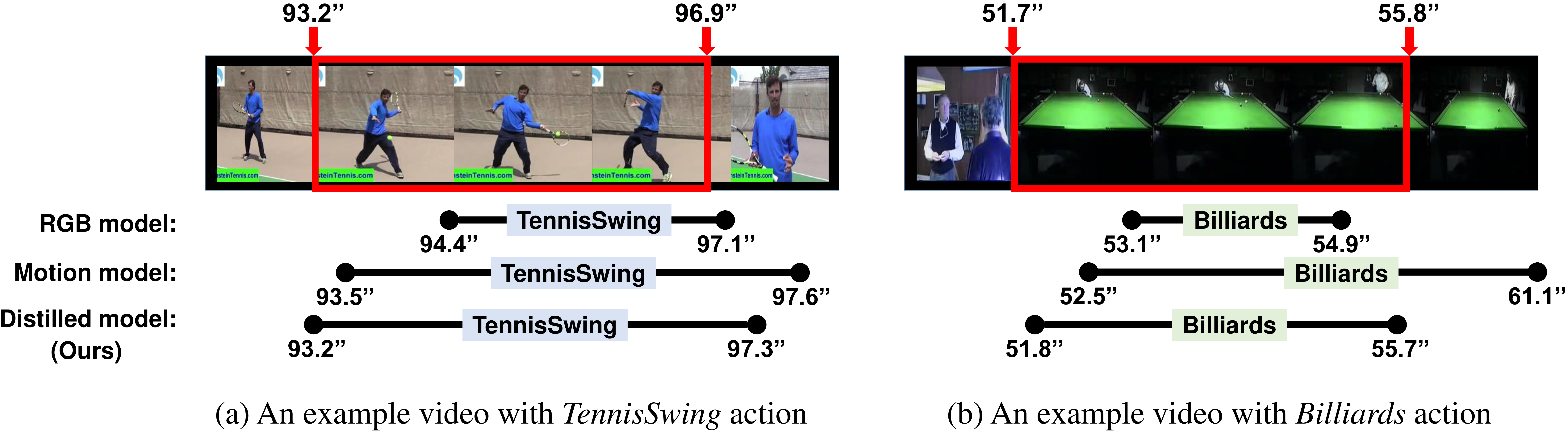}
    \caption{Qualitative results on THUMOS'14.}
    \label{fig:qualitative_results}
\end{figure*}

%% file: tables/experiments_backbones.tex
\begin{table}[t]
\centering
\resizebox{0.96\columnwidth}{!}{
\begin{tabular}{c|c|ccccc|c}
\toprule
\multirow{2}{*}{Backbone} &
\multirow{2}{*}{Distill.} &
\multicolumn{5}{c|}{mAP@IoU (\%)} & \multirow{2}{*}{AVG} \\
& & 0.3 & 0.4 & 0.5 & 0.6 & 0.7 &  \\
      \midrule\midrule
\multirow{3}{*}{TSM18~\cite{lin2019tsm}}
        & \xmark   & 62.3   & 55.2   & 46.2   & 33.8   & 20.4   & 43.6 \\
        & TG   & 64.4   & 58.0   & 49.0  & 37.5  & 24.1  & 46.6 ($+$3.0) \\
        & OF   & 65.3  & 59.5  & 50.9   & 39.6  & 25.5  & 48.2 ($+$4.6) \\
        \midrule
\multirow{3}{*}{TSM50~\cite{lin2019tsm}}
       & \xmark  & 65.0  & 59.2  & 50.0  & 38.2  & 25.0  & 47.5 \\
        & TG   & 68.1  & 61.8   & 52.4   & 41.7   & 27.5   & 50.3 ($+$2.8) \\
        & OF   & 66.5  & 62.3   & 55.3   & 44.5   & 32.9   & 52.3 ($+$4.8) \\
       \midrule
\multirow{3}{*}{I3D~\cite{carreira2017quo}}
       & \xmark  & 53.8   & 47.0   & 38.6  & 30.0  & 19.9  & 37.9 \\
        & TG   & 57.6   & 51.4   & 42.5   & 32.9  & 22.1  & 41.3 ($+$3.4) \\
        & OF   & 57.7  & 52.1  & 44.6 & 34.9 & 24.0 & 42.6 ($+$4.7) \\
       \midrule
\multirow{3}{*}{Slowfast50~\cite{feichtenhofer2019slowfast} }
       & \xmark  & 67.4  & 62.9  & 56.8   & 46.8   & 35.0  & 53.8 \\
        & TG   & 68.9  & 64.1  & 58.1   & 48.2   & 35.6  & 55.0 ($+$1.2) \\
        & OF   & 70.5   & 65.8   & 59.2  & 50.1  & 38.2  & 56.8 ($+$3.0) \\
    \bottomrule
\end{tabular}
}
\caption{
Experiments with different backbones on THUMOS'14. The detection head is fixed to Actionformer~\cite{zhang2022actionformer}. `TG' and `OF' denote temporal gradient and optical flow, respectively.
}
\label{tab:backbones}
\end{table}

%% file: tables/experiments_heads.tex
\begin{table}[t]
\centering
\resizebox{0.98\columnwidth}{!}{
\begin{tabular}{c|c|ccccc|c}
\toprule
\multirow{2}{*}{Head} &
\multirow{2}{*}{Distill.} &
\multicolumn{5}{c|}{mAP@IoU (\%)} & \multirow{2}{*}{AVG} \\
& & 0.3 & 0.4 & 0.5 & 0.6 & 0.7 &  \\
      \midrule\midrule
\multirow{3}{*}{G-TAD~\cite{xu2020g-tad}}
        & \xmark   & 51.4  & 44.7  & 36.0  & 26.4  & 16.8 & 35.1 \\
        & TG   & 54.8  & 48.9  & 38.1  & 28.0  & 18.1  & 37.6 ($+$2.5) \\
        & OF   & 55.3  & 49.4  & 39.2  & 30.6  & 19.7  & 38.8 ($+$3.6) \\
        \midrule
\multirow{3}{*}{TadTR~\cite{liu2022tadtr}}
       & \xmark  & 62.8  & 56.7  & 47.5  & 37.3  & 25.5  & 46.0\\
        & TG   & 63.8  & 57.4  & 49.9  & 39.2  & 26.9  & 47.4 ($+$1.4)\\
        & OF   & 64.1  & 58.3  & 51.2  & 40.9  & 28.8  & 48.7 ($+$2.7)\\
       \midrule
\multirow{3}{*}{Actionformer~\cite{zhang2022actionformer}} 
        & \xmark   & 62.3   & 55.2   & 46.2   & 33.8   & 20.4   & 43.6 \\
        & TG   & 64.4   & 58.0   & 49.0  & 37.5  & 24.1  & 46.6 ($+$3.0) \\
        & OF   & 65.3  & 59.5  & 50.9   & 39.6  & 25.5  & 48.2 ($+$4.6) \\
    \bottomrule
\end{tabular}
}
\caption{
Experiments with different detection heads on THUMOS'14. The backbone is fixed to TSM18~\cite{lin2019tsm}. `TG' and `OF' indicate temporal gradient and optical flow, respectively.
}
\label{tab:heads}
\end{table}

%% file: tables/quantitative_comparison.tex
\begin{table*}
\centering
\resizebox{0.85\textwidth}{!}{
\begin{tabular}{l|c|c|ccccc|c|ccc|c}
\toprule
\multicolumn{1}{c|}{\multirow{2}{*}{Method}} & \multirow{2}{*}{Venue} & \multirow{2}{*}{OF} & \multicolumn{6}{c}{THUMOS'14} &       \multicolumn{4}{|c}{ActivityNet1.3} \\  
&  & & 0.3  & 0.4 & 0.5 & 0.6 & 0.7 & AVG & 0.5   & 0.75   & 0.95   & AVG  \\
\midrule
TAL-Net~\cite{chao2018rethinking} & CVPR'18 &\cmark&53.2&48.5&42.8&33.8&20.8&39.8&38.23&18.30&1.30&20.22\\
BSN\cite{lin2018bsn} & ECCV'18 &\cmark&53.5&45.0&36.9&28.4&20.0&-&46.45&	29.96&8.02&30.03\\
BMN~\cite{Lin2019BMNBN} & ICCV'19 &\cmark&56.0&47.4&38.8&29.7&20.5&38.5&50.07&34.70&8.29&33.85\\
P-GCN~\cite{zeng2019p-gcn} & ICCV'19 &\cmark&63.6&57.8&49.1&-&-&-&48.26&33.16&3.27&31.11\\
G-TAD~\cite{xu2020g-tad} 
 & CVPR'20 &\cmark&54.5&47.6&40.2&30.8&23.4&39.3&50.36& 34.60&9.02 &34.09\\
BC-GNN~\cite{bai2020bcgnn} & ECCV'20 &\cmark&57.1&49.1&	40.4&	31.2&	23.1&	40.2&50.56&34.75&9.37&34.26\\
BU-MR~\cite{zhao2020bottom-up} & ECCV'20 &\cmark&53.9&50.7&45.4&38.0&28.5&	43.3&43.47&33.91&9.21&30.12\\
AFSD~\cite{lin2021afsd} & CVPR'21 &\cmark&67.3& 62.4 &55.5& 43.7 &31.1&52.0&52.38&35.27&6.47&34.39\\
MUSES~\cite{liu2021muses} & CVPR'21 &\cmark&68.9&	64.0&56.9 &46.3&	31.0&	53.4&50.02&34.97&6.57&33.99\\
RTD-Net~\cite{tan2021relaxed} & ICCV'21 &\cmark &68.3&	62.3&	51.9 &	38.8&	23.7&	49.0&47.21&30.68&8.61&30.83\\
VSGN~\cite{zhao2021vsgn} & ICCV'21 &\cmark&66.7&	60.4&	52.4&	41.0&	30.4&	50.2&52.38&36.01&8.37&35.07\\
RCL~\cite{wang2022rcl} & CVPR'22 &\cmark&70.1&62.3&52.9&42.7&30.7&51.7&55.15&39.02&8.27&37.65 \\
RefactorNet~\cite{xia2022refactornet} & CVPR'22 &\cmark&70.7 &65.4&58.6&47.0&32.1&54.8&56.60&40.70 &7.50& 38.60 \\
TAGS~\cite{nag2022proposal-free} & ECCV'22 &\cmark&68.6&63.8&57.0&46.3&31.8&52.8&56.30&36.80&9.60&36.50\\
ReAct~\cite{shi2022react} & ECCV'22 &\cmark&69.2&65.0&57.1&47.8&35.6&55.0&49.60&33.00&8.60&32.60\\
Actionformer~\cite{zhang2022actionformer} & ECCV'22 &\cmark&82.1&77.8&71.0&59.4&43.9&66.8&53.50&36.20&8.20&35.60\\
\midrule
CDC~\cite{shou2017cdc} & CVPR'17 &\xmark& 40.1& 29.4& 23.3& 13.1 &7.9& 22.8& 45.30& 26.00 &0.20 &23.80\\
GTAN\cite{long2019gaussian} & CVPR'19 &\xmark&57.8&	47.2	&38.8&	-&	-&-&52.61&34.14&8.91&34.31\\	
G-TAD*\cite{xu2020g-tad} & CVPR'20 &\xmark&52.5&45.9&37.6&28.5&19.1&36.7&49.22&34.55&4.74&33.17\\
AFSD*\cite{lin2021afsd} & CVPR'21 &\xmark&57.7&52.8&45.4&34.9&22.0&43.6&-&-&-&32.90\\
TadTR*\cite{liu2022tadtr} & TIP'22 &\xmark&59.6&54.5&47.0&37.8&26.5&45.1&49.56&35.24&9.93&34.35\\
E2E-TAD~\cite{liu2022e2e-tad} & CVPR'22 &\xmark&69.4&64.3&56.0&46.4&34.9&54.2&50.47& 35.99 &10.83& 35.10 \\
TAGS$^\dagger$~\cite{nag2022proposal-free} & ECCV'22 &\xmark& 59.8 & 57.2 & 50.7 & 42.6 & 29.1 & 47.9 & 54.44 & 34.95 & 8.71 & 34.95 \\
Actionformer$^\dagger$~\cite{zhang2022actionformer} & ECCV'22 &\xmark&69.8&66.0&58.7&48.3&34.6&55.5& 53.21 & 35.15 & 8.03 & 34.94 \\
Ours & - &\xmark& 70.5  & 65.8  & 59.2  & 50.1  & 38.2  & 56.8  & 53.73 & 35.87  & 8.61  & 35.58   \\
\bottomrule
\end{tabular}
}
\caption{Comparison with state-of-the-art methods. The average mAPs under the IoU thresholds 0.3:0.7 and 0.5:0.95 are reported respectively for THUMOS'14 and ActivityNet1.3.
Entries are grouped by whether the model relies on optical flow (OF) at inference time.
The results of models with asterisk~(*) are taken from Liu~\etal~\cite{liu2022e2e-tad}, while those with dagger~($\dagger$) are the reproduced results by official code.
}
\label{tab:quantitative_comparison}
\end{table*}